\pdfoutput=1

\documentclass[11pt]{article}

\usepackage[]{acl}

\usepackage{times}
\usepackage{latexsym}

\usepackage[T1]{fontenc}

\usepackage[utf8]{inputenc}

\usepackage{microtype}

\usepackage{graphicx}
\usepackage{tabularx}
\usepackage{adjustbox}
\usepackage{booktabs}
\usepackage{multirow}
\usepackage{amsmath}
\usepackage{cuted}
\usepackage{comment}
\usepackage{lipsum}
\usepackage{colortbl}
\usepackage{xcolor}

\title{Relation Extraction with Weighted Contrastive Pre-training \\ on Distant Supervision}

\author{
Zhen Wan  \hspace{1em}
Fei Cheng  \hspace{1em} 
Qianying Liu  \hspace{1em} \\
{\bf Zhuoyuan Mao } \hspace{1em}
{\bf Haiyue Song } \hspace{1em}
{\bf Sadao Kurohashi }\\
Kyoto University, Japan \hspace{1em}
\\
\texttt{\{zhenwan, ying, zhuoyuanmao, song\}@nlp.ist.i.kyoto-u.ac.jp} \\
\texttt{\{feicheng, kuro\}@i.kyoto-u.ac.jp} 
}
\begin{document}
\maketitle
\begin{abstract}

Contrastive pre-training on distant supervision has shown remarkable effectiveness in improving supervised relation extraction tasks. 
However, the existing methods ignore the intrinsic noise of distant supervision during the pre-training stage. 
In this paper, we propose a weighted contrastive learning method by leveraging the supervised data to estimate the reliability of pre-training instances and explicitly reduce the effect of noise.
Experimental results on three supervised datasets demonstrate the advantages of our proposed weighted contrastive learning approach compared to two state-of-the-art non-weighted baselines.Our code and models are available at: \href{https://github.com/YukinoWan/WCL}{https://github.com/YukinoWan/WCL}.
\end{abstract}


\section{Introduction}
Relation extraction (RE) is the task of identifying the relationship between entities mentioned in the text, which can benefit many downstream tasks such as question answering and knowledge base population. Since most of the existing RE models ~\cite{zhang-etal-2020-minimize,Zeng_Zhang_Liu_2020,lin-etal-2020-joint,wang-lu-2020-two,zhong-chen-2021-frustratingly} are trained on the labeled data, the amount of training data limits the performance of supervised RE systems. To tackle this problem, recent work leverage semi-supervised distant supervision (DS) ~\cite{mintz-etal-2009-distant,lin-etal-2016-neural,vashishth-etal-2018-reside,chen-etal-2021-cil} approach to generate abundant training data by aligning knowledge bases (KBs) and raw corpora. However, distantly supervised relation extraction (DSRE) inevitably suffers from wrong labeling noise. Introducing a robust framework that utilizes both the abundant but noisy data from DS and the scarce but accurate data from human annotations becomes a new research line to improve RE systems.

Recent works ~\cite{baldini-soares-etal-2019-matching,DBLP:journals/corr/abs-2102-09681,peng-etal-2020-learning} propose a two-stage RE framework that they first design a RE-oriented task to pre-train BERT on DS data and then fine-tune on human-annotated (HA) datasets.  \citet{peng-etal-2020-learning} use Wikipedia articles as the corpus and Wikidata as the KB in the pre-training stage to construct the DS data, and they introduce a contrastive learning-based method to pre-train BERT on the generated DS data.
Given an anchor instance with a specific relation in the DS data, their contrastive learning method randomly selects one positive sample holding the same relation and maximizes the similarity between the anchor and positive sample.
Meanwhile, their method randomly selects multiple negative samples holding different relations from the anchor and minimizes the similarity between the anchor and negative samples.
The results show that their RE-oriented pre-training can effectively improve the final performance of the RE task on various target datasets.

\begin{figure}[t]
    \centering
    \includegraphics[height=4.0cm]{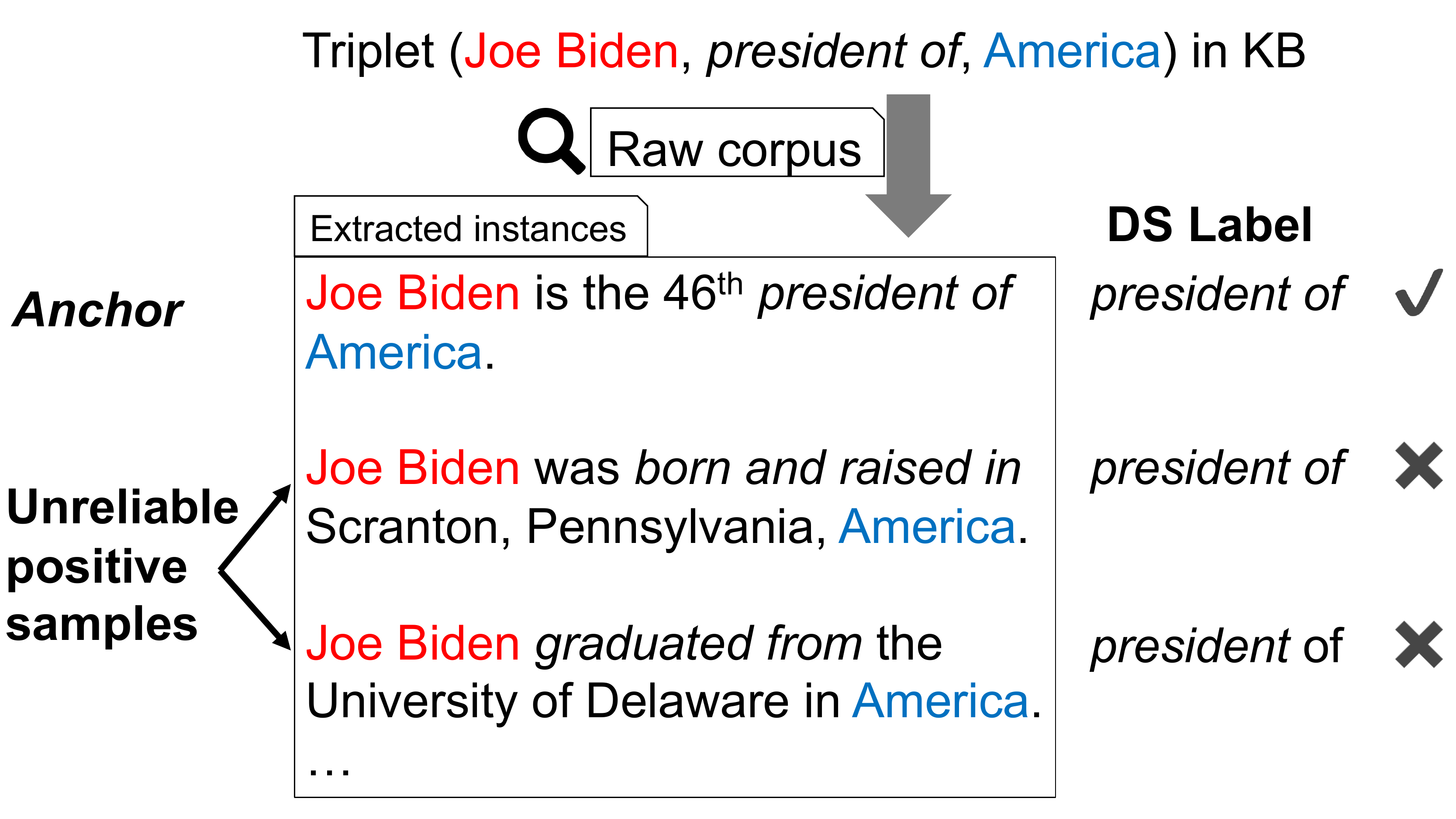}
    \caption{\textbf{An example of unreliable positive samples caused by DS noise}.}
    \label{fig:ds-noise}
\end{figure}

However, in their pre-training stage, they ignore the intrinsic wrong labeling noise in the generated DS data. 
Since their method relies on the DS-labeled relation types to sample positive and negative instances, the noisy labeling problem leads to unreliable samples in Figure~\ref{fig:ds-noise}, potentially limiting the pre-training stage's effectiveness.
To better utilize DS data, we propose a novel weighted contrastive learning framework to both use the abundant DS data and tackle the inevitable DS noise. First, we train a relation classifier on the HA dataset and leverage the classifier to predict the relation type of instances in the DS data. Then for each DS instance, based on the output of the classifier, we can compute the confidence score to measure the reliability of its labeled relation type. Finally, we introduce weights based on computed confidence scores into the contrastive learning loss to focus more on reliable instances while less on noisy ones.

Besides, distant supervision relies on the existing KBs to align raw corpora. To alleviate the need for KBs, we propose a new strategy to extract a triplet set from the HA dataset for generating DS data. 
We also include a KB-derived DS dataset in our experiments to show that our proposal can still work well for regular DS.

In conclusion, we propose a weighted contrastive pre-training approach for supervised relation extraction and introduce its details in Section ~\ref{method}. Then we perform the experiments on three datasets to compare our proposed method with existing baselines in Section ~\ref{experiment}.



\section{Proposed Method} \label{method}

\subsection{Overview}
\begin{figure*}[t]
    \centering
    \includegraphics[height=7.5cm, width=14cm]{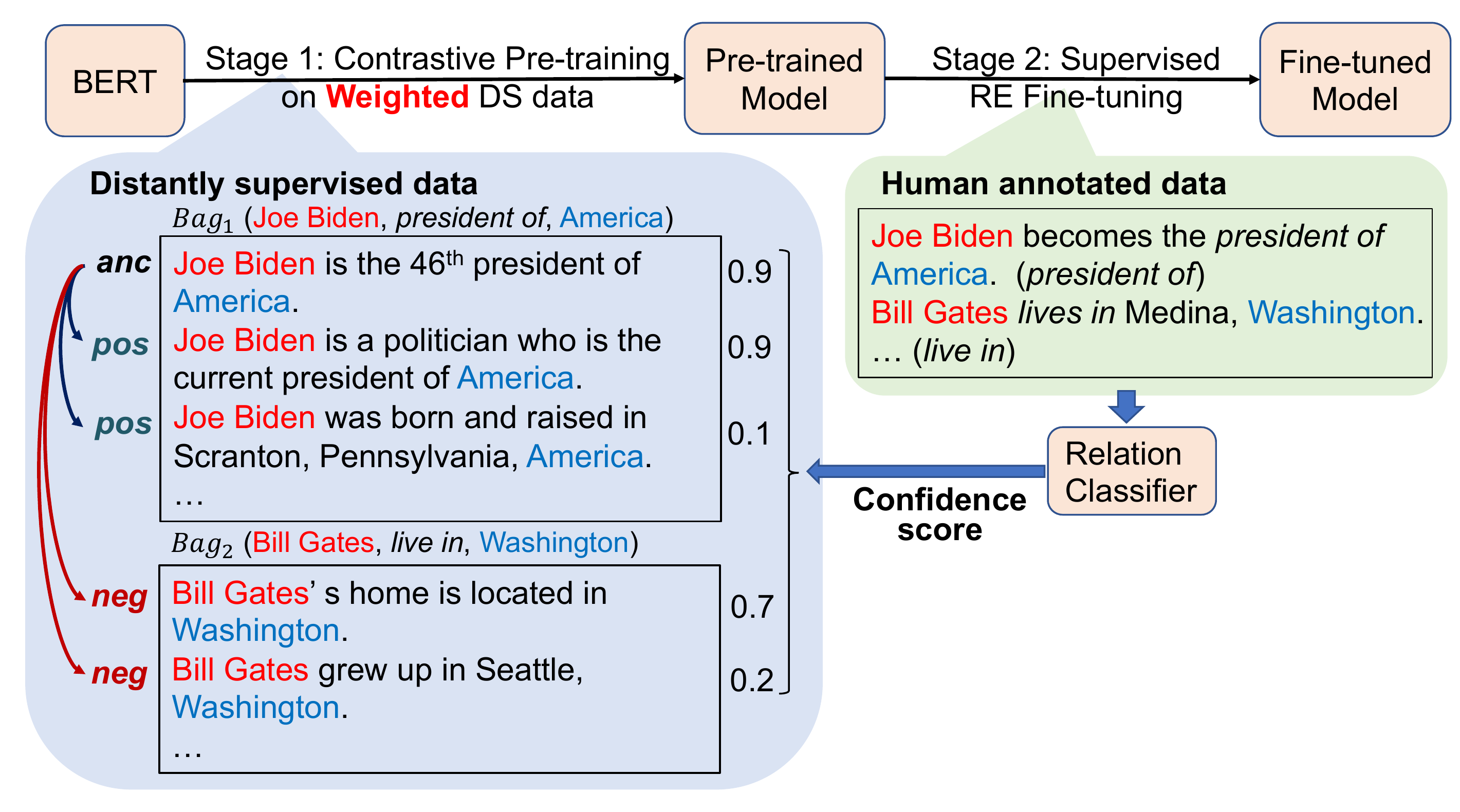}
    \caption{\textbf{Overview of our proposed method}.}
    \label{fig:overview}
\end{figure*}

 We show the overview of our proposal in Figure~\ref{fig:overview}. We start by generating the DS data relying on the HA dataset. Then in the first stage, we introduce a weighted contrastive learning method by leveraging the HA data to estimate the reliability of DS instances for contrastive pre-training. In the second stage, we further fine-tune our pre-trained model on the HA dataset. 

\subsection{Distantly Supervised Dataset Construction}
Since DS uses existing knowledge bases to generate training data, in the case that we have no proper existing KBs in some domains but only the annotated dataset, we first extract all entities based on each sentence, and if any two of them are labeled a relation type, they will generate a triplet with a particular relation. Otherwise, they will still generate a triplet but labeled NA (no relation). After constructing the KB, we can extract sentences containing two entities of each triplet from raw corpora. To balance the number of sentences extracted by each triplet, we also add an upper bound $100$ to the number of extracted sentences.

\subsection{Two-stage RE Framework} \label{framework}
\paragraph{Instance representation}
In our pre-training stage, we use BERT to obtain the representation for each input instance. For the input format, 
we follow PURE \cite{zhong-chen-2021-frustratingly} by adding extra special markers to mark the beginning and the end of two entities. For example, given an instance $x$: ``\textit{\textbf{Joe Biden} is the president of \textbf{America}.}'', the input sequence is ``\textsf{\small [CLS] [H\_CLS]} \textit{Joe Biden} \textsf{\small [H\_SEP]} \textit{is the president of} \textsf{\small [T\_CLS]} \textit{America} \textsf{\small [T\_SEP]}. \textsf{\small [SEP]}''. 
Denote the $k$-th output vector of the BERT encoder as $h_k$.
Assuming $i$ and $j$ are the indices of two beginning entity markers \textsf{\small [H\_CLS]} and \textsf{\small [T\_CLS]}, we define the instance representation as:
\begin{equation}
\mathbf{x}= h_i \oplus h_j
\label{representation}
\end{equation}
where $\oplus$ stands for concatenation. Then we use the instance representation for further reliability estimation and the weighted contrastive learning in the pre-training stage.
\paragraph{Reliability estimation}

With the instance representation, we first fine-tune BERT on the HA dataset as a supervised RE task. Then with the trained relation classifier $\mathcal{F}$, we can make predictions on each instance in the DS data. 
Given an input instance $x$ with DS labeled relation $r$, we can derive the confidence score $c$ to estimate its reliability by:
\begin{equation}
\resizebox{0.5\linewidth}{!}{
$
c = \frac{\exp{(\mathcal{F}(\mathbf{x}, r))}}{\sum_{r'\in R}\exp{(\mathcal{F}(\mathbf{x}, r'))}}
$
}
\label{classifier}
\end{equation}
where $R$ is the set of all relation classes, and $\mathcal{F}(\mathbf{x}, r)$ computes the output of our relation classifier on the labeled class $r$.
Through this approach, we can estimate the reliability of the labeled relation for each DS instance by its corresponding confidence score.
\paragraph{Stage 1: DS weighted contrastive pre-training}
Contrastive learning aims at maximizing the similarity between a given instance and its positive samples while minimizing the similarity between the given instance and its negative samples. 
As for existing work, \citet{peng-etal-2020-learning} focuses on the relationship level that DS instances labeled the same relation are positive samples while DS instances labeled different relations are negative samples. The latest DSRE work ~\cite{chen-etal-2021-cil} augments the anchor as a positive sample to avoid the effect of DS noise. Both works do not explicitly address the problem of unreliable positive and negative samples.

In our work, 
we introduce a robust weighted contrastive learning (WCL) method with the help of reliability estimations for each instance. Given a batch input with multiple bags: ($Batch=\{{B}_{i}\}_{i=1}^G$) where $G$ is the number of bags in one batch, and the labeled relational triplets are different from each other. Each bag $B$ is constructed by a certain relational triplet $({e}_{1}, r, {e}_{2})$ with all instances $x$ inside satisfying this triplet. Moreover, each instance comes along with a confidence score $c$ estimating its reliability: ${B}_{i}=\{{x}_{j}, {c}_{j}\}_{j=1}^{{N}_{i}}$, where ${N}_{i}$ denotes the size of bag ${B}_{i}$. Then the weighted contrastive learning loss of an anchor instance ${x}_{j}$ in the bag ${B}_{i}$ is:
\begin{equation}
\resizebox{\linewidth}{!}{
$
\begin{aligned}
&\mathcal{L}_{WCL}^{(ij)} = -\log \left\{ \sum\limits_{k=1}^{{N}_{i}}c_{j}c_{k}e^{cos(\mathbf{x}_{j}, \mathbf{x}_{k})/\mathcal{T}}/ \right . \\ & \left . ({\sum\limits_{k=1}^{{N}_{i}}c_{j}c_{k}e^{cos(\mathbf{x}_{j}, \mathbf{x}_{k})/\mathcal{T}} + \sum\limits_{{r}_{m}\neq{r}_{j}}c_{j}c_{m}e^{cos(\mathbf{x}_{j}, \mathbf{x}_{m})/\mathcal{T}}}) 
\right\}
\end{aligned}
$
}
\label{clloss}
\end{equation}
where $cos(\cdot)$ denotes the cosine similarity between two instance representations, $\mathbf{x}_{k}$ denotes the representation of a positive instance sampled from the same bag, and ${r}_{m}\neq{r}_{j}$ denotes that negative samples ${x}_{m}$ are selected from all instances in the batch that is labeled a different relation from ${x}_{j}$. We follow ~\cite{NEURIPS2020_d89a66c7} to incorporate multiple positive instances sampled from the same bag. $\mathcal{T}$ denotes a scaling temperature.

With the help of confidence scores, the model will focus on more reliable instances while ignoring unreliable instances, which keep pace with our goal to utilize reliable DS data.

Besides, to inherit the ability of language understanding from BERT and avoid catastrophic forgetting, we also adopt the masked language modeling (MLM) objective from BERT. 

Eventually, we define our final pre-training loss:
\begin{equation}
\resizebox{0.5\linewidth}{!}{
$
\mathcal{L} = \mathcal{L}_{WCL} + \mathcal{L}_{MLM}
$
}
\label{finalloss}
\end{equation}

\begin{table*}[thb]
    \centering
    \resizebox{\linewidth}{!}{
    \setlength{\tabcolsep}{8mm}{
    \begin{tabular}{lrrrrrrrrr}
    \toprule
        \multirow{2}{*}{\textbf{Methods}} & \multicolumn{2}{c}{\textbf{i2b2 2010VA}} & \multicolumn{2}{c}{\textbf{ACE05}}  & \multicolumn{2}{c}{\textbf{Wiki20m}} \\
        &25\%&100\%&25\%&100\%&25\%&100\%\\
        \toprule
        FT & 66.86&	75.22&	\textbf{62.81}&	\textbf{70.41}& 68.87&88.54\\
        \hline
        CIL + FT & 67.92&	75.39	&59.72& 69.69	&89.67	&	91.64\\
        RECN + FT & 67.65&	75.43	&60.34& 69.40	&	89.23&	91.96\\
        WCL + FT (ours) & \textbf{68.50}& \textbf{76.15}	&61.30 & 69.47	&\textbf{90.28}&\textbf{92.67}\\
        
        \bottomrule
    \end{tabular}}
    }
    \caption{\textbf{Evaluation results on various datasets}. 25\% denotes the low-resource setting, and 100\% denotes the full-resource setting. We compute three-run average Micro-F1 for our proposed methods in all the results.}
    \label{results}
\end{table*}

\paragraph{Stage 2: Supervised relation extraction}
We then fine-tune the pre-trained model on HA datasets with state-of-the-art (SOTA) methods. For i2b2 2010VA, we follow BLUEBERT~\cite{peng-etal-2019-transfer} by treating the relation extraction task as a sentence classification and replacing two named entities in the sentence with predefined tags. For the other two datasets, we follow the encoding method of PURE~\cite{zhong-chen-2021-frustratingly} as introduced at the beginning of Section~\ref{framework}.

\begin{table}[t]
    \centering
    \resizebox{\linewidth}{!}{
    \begin{tabular}{lrrrr}
    \toprule
        Dataset & \# Rel. & \# Train & \# Dev & \# Test \\
        \hline
        i2b2 2010VA & 8 & 3,120 & 11 & 6,147 \\
        ACE05 & 6 & 10,051 & 2,424 & 2,050 \\
        Wiki20m & 80 & 8,279 & 4,140 & 28,977 \\
        \hline
        ACE05 (NP) & 6 & 3,939 & 922 & 923 \\

        \bottomrule
    \end{tabular}
    }
    \caption{\textbf{Statistics of datasets}. Rel. denotes relation types. NP denotes removing pronoun from ACE05.}
    \label{annotated data}
\end{table}

\begin{table}[t]
    \centering
    \resizebox{\linewidth}{!}{
    \begin{tabular}{lrrrr}
    \toprule
        Dataset & \# Triplets & Corpora & \# DS Ins. (NA) \\
        \hline
        i2b2 2010VA & 2,777 & MIMIC-III & 36K (76K) \\
        ACE05  & 3,883 & Gigaword5 & 98K (461K) \\
        Wiki20m  &-& Wiki20m & 286K (698K)  \\
        \hline
        ACE05 (NP)  & 3,218& Gigaword5 & 60K (273K) \\

        \bottomrule
    \end{tabular}
    }
    \caption{\textbf{Statistics of DS data}. Triplets are extracted from the HA dataset. DS Ins. denotes relational instances generated by DS. NA denotes the no-relation instances. NP denotes removing pronoun from ACE05.}
    \label{ds data}
\end{table}

\begin{table}[thb]
    \centering
    \resizebox{\linewidth}{!}{
    \begin{tabular}{lrrrrrrr}
    \toprule
        \multirow{2}{*}{\textbf{Methods}} & \multicolumn{2}{c}{\textbf{ACE05}}  & \multicolumn{2}{c}{\textbf{Wiki20m}} \\
        &25\%&100\%&25\%&100\%\\
        \toprule
        FT & 	\textbf{60.45}&	\textbf{69.82}& 66.58&89.38\\
        \hline
        CIL + FT 	&60.12& 69.36	&90.25	&	91.26\\
        RECN + FT 	&58.68& 68.04	&	90.18&	91.75\\
        WCL + FT (ours) 	&60.20 & 69.73	&\textbf{91.06}&\textbf{92.94}\\
        
        \bottomrule
    \end{tabular}}
    \caption{\textbf{Evaluation results on the development set datasets}. 25\% denotes the low-resource setting, and 100\% denotes the full-resource setting.}
    \label{dev result}
\end{table}


\section{Experiments} \label{experiment}
\subsection{Setup}
\paragraph{HA and DS datasets}
We evaluate our approach on three HA relation extraction datasets: i2b2 2010VA, ACE05, and Wiki20m. Table~\ref{annotated data} shows the statistics of each dataset. The i2b2 2010VA is a medical domain RE dataset, while the other two datasets are collected from general domains. We generate the DS data for i2b2 2010VA and ACE05 from corresponding raw corpora. Meanwhile, Wiki20m is a regular KB-based distantly supervised RE dataset containing both DS data and HA data and it is worth noting that we intend to show that our method can also work well on existing DS datasets. Table~\ref{ds data} shows the statistics of DS data.


\paragraph{Baselines}

We have a naive baseline by directly fine-tuning (FT) on each dataset as a supervised RE task. We set two two-stage framework baselines: the first one is to use the SOTA method RE-Context-or-Names (RECN)~\cite{peng-etal-2020-learning} in pre-training, and the second one is to use the SOTA DSRE method Contrastive Instance Learning (CIL)~\cite{chen-etal-2021-cil} in pre-training.

\paragraph{Implementation details}


To further confirm the effectiveness of our proposal, we also conduct the experiments in the low-resource setting by randomly selecting 25\% of the full HA data to construct the DS data for pre-training and finally fine-tune on this 25\% HA data. Refer to Appendix~\ref{imple} for other implementation details.

\subsection{Main Results}
Table~\ref{results} compares our model to other baselines. From the results, we can observe that: (1) For both the i2b2 2010VA and the Wiki20m, all two-stage models outperform the FT baseline, which indicates the effectiveness of our strategy to construct DS data from HA datasets, especially in the low-resource setting. (2) For both the i2b2 2010VA and the Wiki20m, our proposed model achieves the best F1 scores over all baselines. 
This improvement shows that it is worth estimating the reliability of each DS instance with the help of HA datasets in our weighted contrastive pre-training. 
(3) For the ACE05,
the pre-training methods cannot outperform the FT baseline.
To analyze this problem, we perform extra experiments on ACE05 in Section~\ref{ace05}.

Besides, we also compare performances on the development set of two datasets as shown in Table~\ref{dev result}, the experiment results emphasize the consistent improvement of our proposed methods.\color{black}

\subsection{Further Analysis}
\label{ace05}
\begin{table}
    \centering
    \resizebox{0.8\linewidth}{!}{
    \begin{tabular}{lrrrrrrrrr}
    \toprule
        \multirow{2}{*}{\textbf{Methods}}  & \multicolumn{2}{c}{\textbf{ACE05 (no pronouns)}} \\
        &25\%&100\%\\
        \toprule
        FT & 62.22&	70.29&	\\
        \hline
        CIL + FT & 63.31&69.76	\\
        RECN + FT & 62.43&70.09	\\
        WCL + FT (ours) & \textbf{64.45}& \textbf{71.10} \\
        
        \bottomrule
    \end{tabular}
    }
    \caption{\textbf{Evaluation on ACE05 after removing pronouns}.}
    \label{npresult}
\end{table}
We find that ACE05 contains many pronoun entities, for example, \textit{"\textbf{He} lives in \textbf{America}."}. As pronoun entities such as \textit{"\textbf{He}"} naturally come along with much more severe noise in DS, we also conduct extra experiments by removing sentences containing pronoun entities in ACE05 and the corresponding DS data to confirm the effect of pronouns.

After removing pronoun entities in ACE05, as shown in Table~\ref{npresult}, our model outperforms all baselines, including FT, which indicates that pronoun entities bring mishandled noise in the pre-training stage and limit the effect of our DS data construction approach.

\section{Conclusions}
We introduce a weighted contrastive pre-training method by leveraging the HA dataset to estimate the reliability of instances in the abundant DS data. 
To alleviate the need for KBs, we also propose to construct DS data based on the triplets derived from the HA dataset for pre-training.
Experimental results demonstrate that our proposed method outperforms SOTA work on target HA datasets. 


\section*{Limitations}
In this paper, we propose a weighted contrastive pre-training approach for supervised relation extraction. 

While our approach is simple and effective, one limitation is that the reliability estimation requires a certain amount of annotated data. Under certain settings such as few-shot learning, large-scale labeled data may not be available, and the reliability of DS data could be estimated in an unsupervised manner based on similarity-based metrics, which we leave as future work.


Another limitation is that the distant supervision of ACE05 contains additional noise caused by pronoun entities. In Section~3.3, we do the investigation by temporally removing them. In future work, we assume more strict DS extraction criteria (e.g. only entity pairs located in the same clause) might reduce the production of such noise and alleviate this situation. 

\section*{Acknowledgements}
This work was partially supported by MHLW PRISM Grant Number 21AC5001, JSPS KAKENHI Grant Number 22J1371, JSPS KAKENHI Grant Number 21J23124, and JSPS KAKENHI Grant Number 21H00308.
\bibliography{anthology,custom}
\bibliographystyle{acl_natbib}

\appendix
\section{Implementation Details}
\label{imple}
During the construction of DS data, we use the preprocessing tool NLTK to split raw corpora into sentences.

We use \textit{bert-base-uncased}~\cite{devlin-etal-2019-bert} as the base encoders for ACE05, ACE05 (no pronouns), and Wiki20m, for a fair comparison with previous works. We also use \textit{bluebert}~\cite{peng-etal-2019-transfer} as the base encoder for i2b2 2010VA since the SOTA performance is achieved based on this effective medical domain BERT. 

For baseline models, we modify their official implementations to fit our experiments and follow the model settings in their papers. 
For our proposed method, the primary hyperparameters in the experiments are batch size, bag size, and contrastive learning temperature that directly influence the weighted contrastive learning loss, and we show our searching ranges and best values in Table ~\ref{hyperparameters}. 

We used 8 NVIDIA A100 for pre-training and 2 NVIDIA RTX3090 for fine-tuning.

\begin{table}[t]
    \centering
    \resizebox{0.8\linewidth}{!}{
    \begin{tabular}{lrrrr}
    \toprule
        Hyperparameter & Range & Best \\
        \hline
        Bag size& 2-8 & 4  \\
        Batch size & 8-32 & 16\\
        Temperature  &0.05-1.0& 0.2  \\

        \bottomrule
    \end{tabular}
    }
    \caption{Hyperparamter optimazition.}
    \label{hyperparameters}
\end{table}

\end{document}